\documentclass{article}
\usepackage{spconf,amsmath,amssymb,graphicx,url}
\usepackage{multirow}
\usepackage{hyperref}
\usepackage{stfloats}

\title{Unsupervised Defect Detection for Surgical Instruments}

\name{%
\begin{tabular}{@{}c@{}}
Joseph Huang,  
Yichi Zhang,
Xiaoyu Ji,
Jingxi Yu, 
Wei Chen,
Seunghyun Hwang, \\ 
Qiang Qiu,
Amy R. Reibman,
Edward J. Delp,
Fengqing Zhu
\end{tabular}}
\address{Purdue University\\
    School of Electrical and Computer Engineering\\
    465 Northwestern Ave, West Lafayette, IN 47907}

\begin{document}
%
\maketitle
\begin{abstract}
Ensuring the safety of surgical instruments requires reliable detection of visual defects. However, manual inspection is prone to error, and existing automated defect detection methods, typically trained on natural/industrial images, fail to transfer effectively to the surgical domain. We demonstrate that simply applying or fine-tuning these approaches leads to issues: false positive detections arising from textured backgrounds, poor sensitivity to small, subtle defects, and inadequate capture of instrument-specific features due to domain shift. To address these challenges, we propose a versatile method that adapts unsupervised defect detection methods specifically for surgical instruments. By integrating background masking, a patch-based analysis strategy, and efficient domain adaptation, our method overcomes these limitations, enabling the reliable detection of fine-grained defects in surgical instrument imagery.
\end{abstract}

\section{Introduction}
\label{sec:intro}

The cleanliness and structural integrity of surgical instruments is paramount for patient safety and successful surgical outcomes~\cite{rutala2019guideline}. Currently, most clinical environments rely on manual visual inspection by trained technicians to identify defects such as contamination, rust, or physical damage. However, this process is labor-intensive, subjective, and prone to human error, particularly when defects are subtle or rare. Due to the large volume and diverse types of surgical tools, the limitations of manual inspection are increasingly pronounced, highlighting the urgent need for scalable, automated quality assurance solutions.

\begin{figure}[ht]
    \centering
    \includegraphics[width=0.42\textwidth]{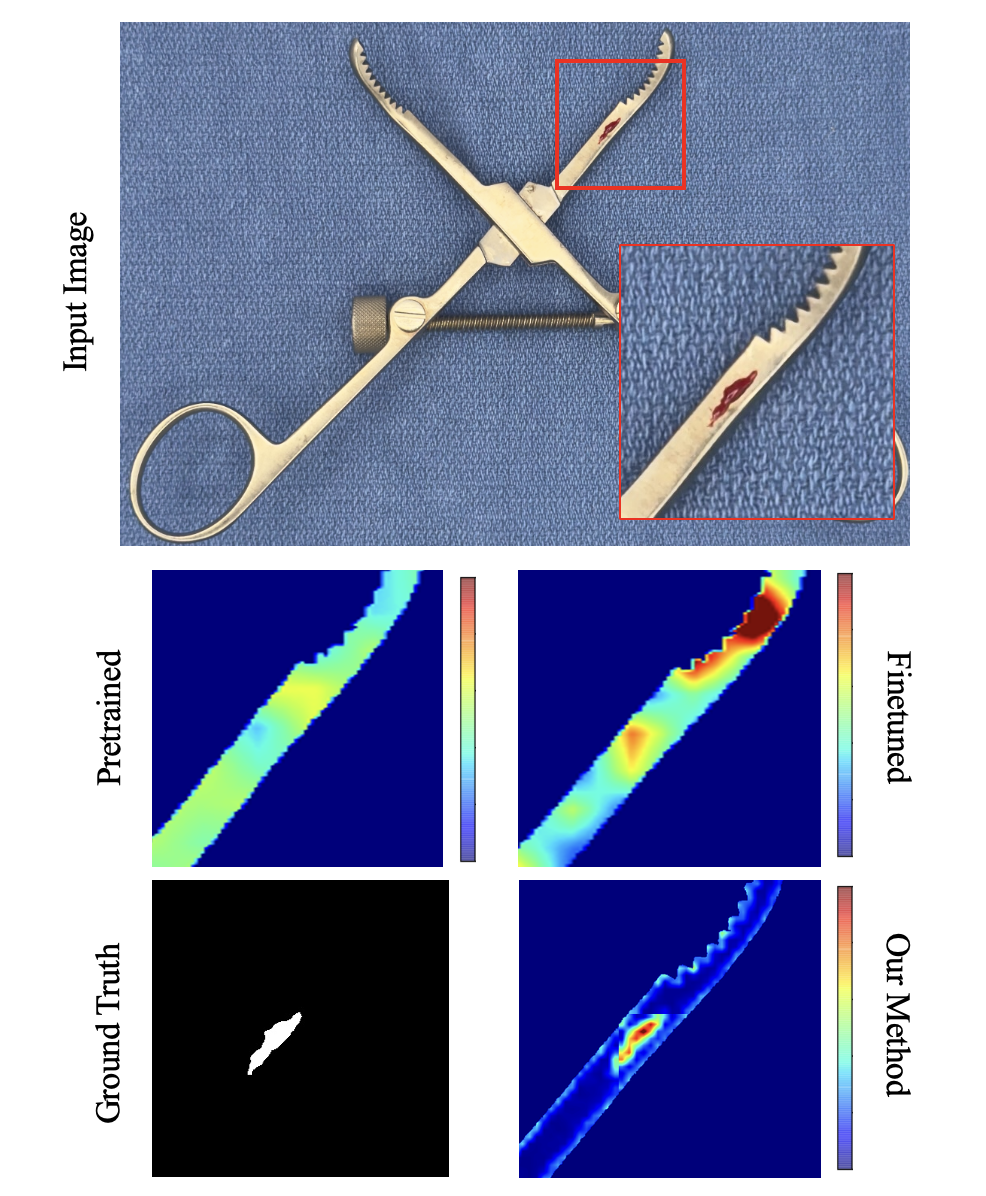}
    \caption{Example of a surgical instrument image with a simulated defect, shown alongside heatmap results from the pretrained Dinomaly, finetuned Dinomaly, and our proposed method. Red regions highlight areas more likely to contain defects, while blue regions indicate areas more likely to be normal.}
    \label{fig:intro_fig}
\end{figure}

Unsupervised defect detection (also known as anomaly detection) presents a promising avenue for automation~\cite{liu2024deep,jiang2022softpatch,patchcore,guo2025dinomaly}. Unlike supervised approaches~\cite{pang2021deep}, these methods model the distribution of normal (defect-free) data and flag deviations as anomalies, thereby eliminating the need for large, annotated defect datasets. This is especially critical in medical contexts, where collecting comprehensive labeled defect data is resource-intensive and where defective instruments are commonly classified as bio-hazards. Recent advances in defect detection~\cite{patchcore,guo2025dinomaly} have demonstrated remarkable performance in industrial inspection tasks, achieving superior results on benchmarks such as the MVTec dataset~\cite{mvtec,heckler2025mvtec}.

Nevertheless, applying these advanced methods directly to surgical instrument inspection reveals a substantial performance gap, as demonstrated in this paper. The unique characteristics of the surgical instrument domain necessitate tailored adaptations. We identify three key obstacles that hinder the performance of existing methods:
\textbf{(1) Background Interference:} The variable backgrounds common in surgical inspection settings (e.g., textured surgical towels or sterilization wraps) introduce spurious signals, resulting in high false-positive rates.
\textbf{(2) Subtle Defects:} Clinically significant defects—such as fine scratches, small cracks, or minor contamination—are often subtle and easily missed by models tuned for more conspicuous industrial defects.
\textbf{(3) Domain Shift:} Pretrained encoders utilized in unsupervised anomaly detection, such as Vision Transformers (ViT)~\cite{dosovitskiy2020image,oquab2023dinov2}, are typically optimized on natural images (e.g., ImageNet 1k~\cite{deng2009imagenet}). They struggle to capture the specialized features of surgical instruments (e.g., metallic surfaces, diverse geometries) necessary for accurate inspection.

In this work, we introduce a method designed to bridge this gap and enable robust surgical instrument inspection by adapting existing unsupervised methods. Specifically, we propose a pre-processing strategy to suppress background-induced false positives, facilitating the image analysis to focus exclusively on the instrument regions. We also introduce a patch-based process, enabling higher resolution analysis and improving sensitivity to small, fine-grained defects. Finally, we mitigate domain shift through Low-Rank Adaptation to pretrained encoders, enhancing their ability to represent instrument-specific features efficiently. To support evaluation, we constructed a dataset that contains defect-free surgical instrument images and realistic instrument defects. 

Through these integrated strategies, we demonstrate that methods such as Dinomaly~\cite{guo2025dinomaly} and DRAEM~\cite{Zavrtanik_2021_ICCV} can be effectively deployed for surgical instrument inspection, transforming them into viable and reliable tools for clinical quality assurance.

\begin{figure*}[htbp]
    \centering
    \includegraphics[width=0.91\textwidth]{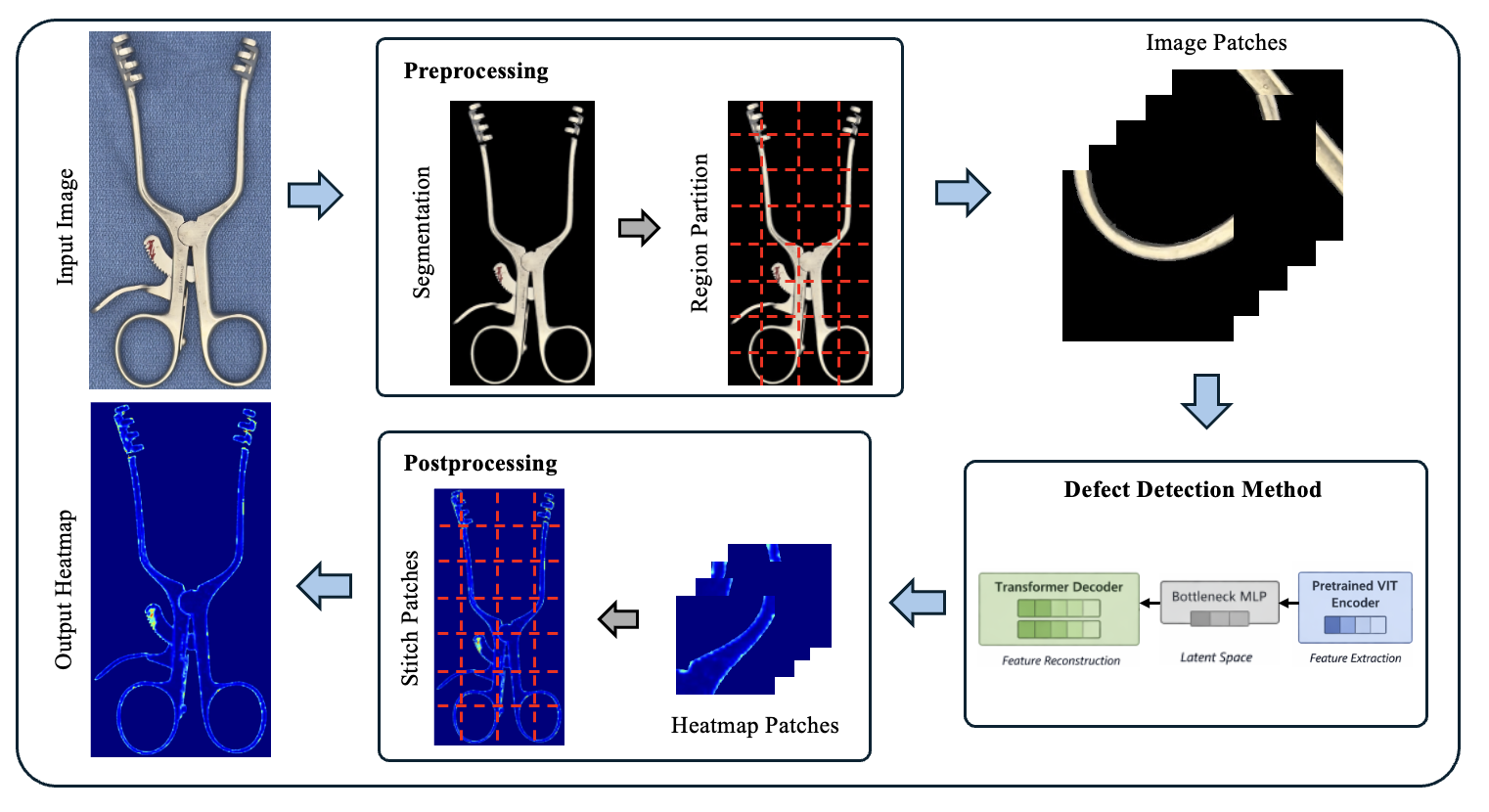}
    \caption{Overview of the proposed defect detection method for surgical instruments. The method consists of three main parts: (1) Preprocessing: the input image undergoes background segmentation and is partitioned into uniform patches; (2) Baseline Detection Method: these patches are fed into a defect detection method, \textit{i.e.}, Dinomaly~\cite{guo2025dinomaly}. If the method is transformer based, LoRA adaption is implemented; (3) Postprocessing: the resulting reconstruction error maps are stitched back together to form a complete output heatmap, localizing the specific defect on the instrument.}
    \label{fig:overview}
\end{figure*}


\section{Proposed Method}
\label{sec:methodology}

Our approach focuses on adapting unsupervised defect detection methods to the unique challenges posed by surgical instrument inspection. The method is designed to be model-agnostic and can be integrated with various baseline models. We address the primary challenges: background interference, subtle defects, and domain shift through an integrated approach combining segmentation, region partition, and low-rank domain adaptation as shown in Fig.~\ref{fig:overview}.

\subsection{Background Masking}
In hospital inspection settings, surgical instruments are often placed over colored and textured backgrounds, such as surgical towels. These background textures frequently generate false defect signals, making detection less accurate. To ensure the model focuses exclusively on the instrument, we introduce a pre-processing step that leverages the Segment Anything Model (SAM)~\cite{kirillov2023segment} for precise segmentation of the instrument from its background.

SAM is applied in fully automatic mask generation mode, without manual prompts or annotations. Specifically, we use SAM’s automatic mask generator, which densely samples a predefined grid of point prompts across the image to produce a set of candidate segmentation masks. Among these candidates, the mask with the largest pixel area is selected, under the assumption that the surgical instrument constitutes the dominant object in the scene.

For computational efficiency during segmentation, images are spatially downsized so that their width is $512$ pixels while preserving the original aspect ratio. The downsized image is processed by SAM, generating a segmentation mask. This mask is subsequently upscaled to the original image dimensions and applied to the high-resolution input image, effectively isolating the instrument. All pixels outside the instrument mask are zeroed out.

By removing background clutter, the model avoids being influenced by irrelevant artifacts. This focused analysis enhances sensitivity to true defects on the instrument surface while reducing false positives caused by background variations. Overall, this preprocessing step helps to significantly improve the reliability and robustness of defect detection.

\subsection{Patch-based Strategy}
Small and subtle defects, such as fine scratches or minor contamination, are easily overlooked when analyzing the entire image. This is often because the image is down-sized to fit model input constraints, resulting in a loss of information critical for identifying these defects.

To improve sensitivity to these fine-grained defects, we employ a patch-based analysis strategy. Each high-resolution image (after background masking) is divided into non-overlapping patches of size $256\times256$. Defect detection is then performed at the patch level. This approach allows the model to analyze local details at a higher resolution.

Importantly, patching increases the relative visibility of small defects. For example, a very small blood stain could occupy only 10 pixels in a full image (3024$\times$4032) equating to $8.2 \times 10^{-5}\%$ of the whole image. When this image is divided into patches, that same defect would then occupy 0.015\% of the patch. Thus, by dividing the entire image into patches, we are able to effectively boost the signal-to-noise ratio and make the defect easier for the model to detect.

After processing, patch-level defect scores or heatmaps are reconstructed into a full-image defect map by stitching the patches according to their original spatial locations. To mitigate isolated noisy responses, the resulting defect map is thresholded using a predefined threshold which maximizes the pixel level AUROC~\cite{hanley1982meaning}. Furthermore, the original SAM-generated segmentation mask is reapplied to the final defect map to ensure that any residual noise outside the instrument region is eliminated. This strategy improves both the localization accuracy and the sensitivity to clinically meaningful small defects.

\subsection{Low-Rank Adaptation for Domain Shift}
Unsupervised defect detection methods often rely on encoders pretrained on large-scale natural image datasets. The significant distributional difference between natural images and surgical instruments results in a domain gap, causing a loss of fine-grained, instrument-specific features critical for accurate inspection. To address this, we employ Low-Rank Adaptation (LoRA)~\cite{hu2022lora} to fine-tune the pretrained encoders efficiently on the target domain (nominal surgical instruments). LoRA introduces trainable low-rank matrices into the existing layers of the network, allowing the model to adapt without modifying the original pretrained weights.

In our implementation, the parameters of the pretrained encoder are frozen. Trainable low-rank matrices are inserted specifically into the self-attention layers, targeting the query–key–value (\textit{qkv}) projections and the output projection (\textit{proj}) modules. Given a frozen weight matrix $W_0 \in \mathbb{R}^{d \times k}$, LoRA reparameterizes the adapted weight as
\begin{equation}
W = W_0 + \Delta W, \quad \Delta W = BA,
\end{equation}
where $A \in \mathbb{R}^{r \times k}$ and $B \in \mathbb{R}^{d \times r}$ are learnable matrices of rank $r \ll \min(d,k)$. Here, $d$ is the output dimension of the linear layer (e.g., the embedding dimension in a ViT attention projection), and $k$ is the input dimension of the layer (e.g., also the embedding dimension for $qkv$ projections). Only $A$ and $B$ are optimized during training, while $W_0$ remains fixed. This reduces the number of trainable parameters from $\mathcal{O}(dk)$ to $\mathcal{O}(r(d+k))$, with $r$ typically much smaller than $d$ or $k$. 

We apply LoRA to the self-attention projections because these layers govern how local patch features are globally aggregated in Vision Transformers. Adapting the \textit{qkv} and output projection layers allows the model to refine attention patterns for capturing instrument-specific structures and surface textures, while earlier convolutional or patch embedding layers remain unchanged to preserve general visual features. The rank $r$ is set to 8, providing sufficient adaptation capacity for instrument-specific variations while maintaining parameter efficiency and avoiding overfitting.

LoRA makes the adaptation process computationally efficient and less prone to overfitting. By utilizing LoRA, the encoder retains its general visual knowledge while adapting to the subtle, instrument-specific patterns crucial for detecting defects~\cite{biderman2024lora}.

Collectively, these adaptations enhance the effectiveness of existing defect detection methods for surgical instrument inspection. By combining background masking, patch-based analysis, and low-rank adaptation, the approach addresses the main challenges in this domain while remaining compatible with a wide range of base models~\cite{guo2025dinomaly,liang2023omni,jiang2022softpatch,patchcore,he2024mambaad}.

\section{Experiments}
\label{sec:experiments}

\begin{figure*}[t]
    \centering
    \includegraphics[width=1.0\textwidth]{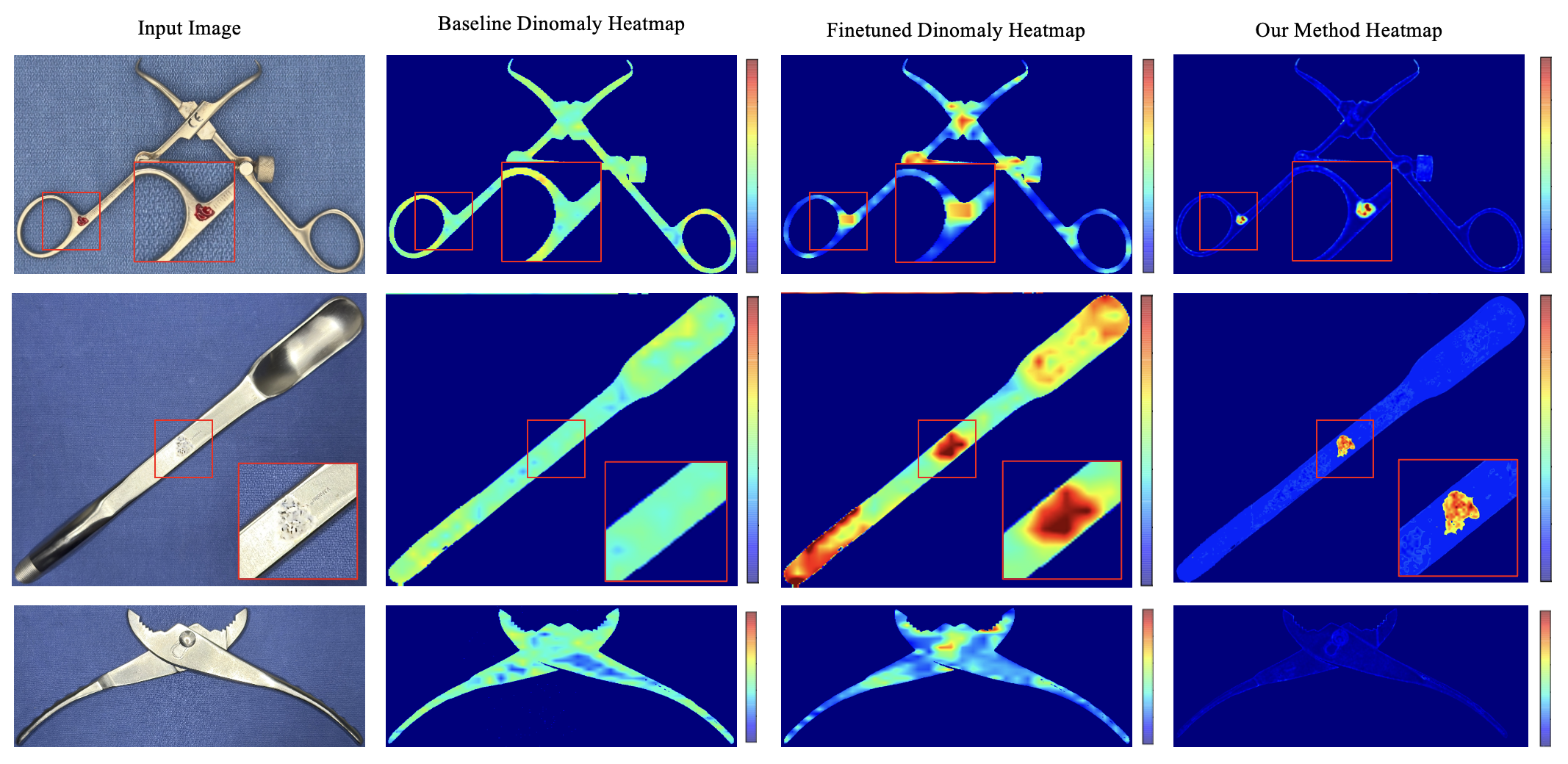}
    \caption{Visualization of image in our surgical instrument dataset using Dinomaly as the base method. Top row: example surgical instrument image including simulated blood stain. Middle row: example surgical instrument image including a pore defect. Bottom row: example surgical instrument image with no defects. Red regions highlight areas the model believes are more likely to contain defects, while blue regions indicate areas more likely to be normal.}
    \label{fig:full_images}
\end{figure*}

\subsection{Experimental Setup}
\textbf{Dataset:} We constructed a surgical instrument dataset containing both normal (clean) and defective examples. To ensure consistent illumination, instruments were photographed inside a lightbox. All images were captured using an iPhone 16 Pro with the ProCamera app under controlled settings: 24mm focal length, 3024 × 4032 pixels, ƒ/1.78 aperture, auto focus, manual white balance (5600K color temperature, 0 tint), ISO 50, shutter speed 1/1005 s, and 0 EV.

The dataset consists of 20 classes of surgical instruments, each represented by 10-20 sample images without defects (340 total). These images were used for training. 



Collecting a large-scale dataset of real defective surgical instruments is inherently challenging. In clinical practice, defective instruments are rare, and those identified are typically classified as bio-hazardous, making them unsuitable for safe imaging or data collection. To address this limitation while ensuring rigorous evaluation, we constructed a test set of 300 defective samples using simulated defects that replicate the primary failure modes of surgical instruments.

For 15 of these samples, we applied dyes (food coloring) to the instrument surfaces. This approach reproduces the irregular staining patterns of dried blood and oxidation, which are visually distinct from the instruments’ metallic specular reflections, providing a physically realistic and easily observable type of defect.

The remaining 285 defective samples were generated using the TF-IDG (Training-Free Industrial Defect Generation) \cite{xu2025training}. Unlike simple Gaussian noise or random masking, TF-IDG leverages diffusion models to create coherent, localized defects that conform to the instrument’s geometry while deviating from its expected texture. Using this approach, we simulated corrosion, cracks, cuts, and pores, enabling rigorous evaluation on complex, structurally realistic defects.

\textbf{Baselines and Implementation:} We evaluated two distinct unsupervised defect detection methods. Dinomaly~\cite{guo2025dinomaly} is a feature-based approach utilizing a Vision Transformer (ViT) pretrained on natural images. DRAEM~\cite{Zavrtanik_2021_ICCV} is a reconstruction-based method that employs a denoising autoencoder trained on images with synthetic defects.

We evaluated the models under three conditions: (1) Using the officially released, pretrained weights (``Pretrained"); (2) Fully fine-tuning the models on our normal surgical instrument dataset (``Finetuned"); and (3) Integrating the models into our proposed method, including masking, patching, and LoRA adaptation (``+ Our Method").

During training/adaptation, all methods were restricted to normal images. Thresholds for defect scoring were determined by maximizing the F1 score on a separate validation split.

\textbf{Evaluation Metrics:} We report the Area Under the Receiver Operating Characteristic curve (AUROC)~\cite{hanley1982meaning} at the pixel level (P-AUROC), following the standard evaluation protocol established by the MVTec AD benchmark~\cite{mvtec,heckler2025mvtec}.



\subsection{Main Results}
Table~\ref{tab:main-table} presents the pixel-level AUROC (P-AUROC) comparison of Dinomaly and DRAEM across three experimental settings: direct use of pretrained models (Pretrained), full fine-tuning on the target domain (Finetuned), and integration with our proposed method (+ Our Method).

Overall, directly applying pretrained models yields suboptimal performance across all defect types, indicating limited generalization to the surgical instrument domain. Full fine-tuning improves performance substantially for both Dinomaly and DRAEM. However, the results remain inconsistent across defect categories, with several defect types still exhibiting weak P-AUROC scores.

In contrast, incorporating the baseline models into our proposed method consistently improves performance across nearly all defect types. For Dinomaly, our method increases P-AUROC for all five defect categories, achieving particularly strong gains for blood, pores, and cracks. For DRAEM, our method yields large improvements for blood, cut, corrosion, and crack defects, while providing more modest gains for pores. These results demonstrate that our method enhances both texture-based and structural defect localization, even when the underlying backbone architectures differ.

Importantly, these improvements are achieved with minimal additional overhead. Our method preserves the original inference pipeline of each baseline model and introduces no additional runtime cost during deployment. LoRA adds less than $1\%$ trainable parameters and is applied only to the transformer-based Dinomaly model, while SAM-based masking is a one-time preprocessing step per session. Note that LoRA is not applied to DRAEM due to its CNN-based architecture.

Finally, we intentionally exclude comparisons with fully supervised defect detection methods. Such approaches rely on dense pixel-level annotations or large collections of labeled defective samples, which are impractical in real-world surgical inspection settings. Acquiring such annotations would require significant involvement from trained clinical personnel, whose time is limited and prioritized for clinical duties, making manual labeling of rare defects impractical.

Figure \ref{fig:full_images} provides a visualization of the results, illustrating the ability of our method to accurately localize defects on the instrument while suppressing background noise.

\begin{table}[ht]
\centering
\resizebox{\columnwidth}{!}{%
    \begin{tabular}{llccc}
    \hline
    Method & Defect & Pretrained & Finetuned & + Our Method \\
    \hline
    \multirow{2}{*}{Dinomaly~\cite{guo2025dinomaly}} 
     & Blood & 0.802 & 0.898 & 0.954 \\
     & Pores & 0.480 & 0.883 & 0.927 \\
     & Cut & 0.523 & 0.862 & 0.902 \\
     & Corrosion & 0.429 & 0.938 & 0.942 \\
     & Crack & 0.549 & 0.866 & 0.933 \\

    \hline
    \multirow{2}{*}{DRAEM~\cite{Zavrtanik_2021_ICCV}} 
     & Blood & 0.530 & 0.792 & 0.966 \\
     & Pores & 0.649 & 0.753 & 0.823 \\
     & Cut & 0.410 & 0.853 & 0.898 \\
     & Corrosion & 0.430 & 0.919 & 0.928 \\
     & Crack & 0.467 & 0.755 & 0.892 \\
    \hline
    \end{tabular}
}
\caption{Main results comparing different methods with and without our method. Each entry shows \textbf{P-AUROC}.}
\label{tab:main-table}
\end{table}

\subsection{Ablation Study}
We performed an ablation study on Dinomaly to quantify the impact of each component in our method. Table~\ref{tab:ablation} reports the performance across the following configurations: (1) the Dinomaly baseline, (2) Dinomaly with fine-tuning, (3) Dinomaly with masking, (4) Dinomaly with masking and patching, and (5) Dinomaly with low-rank adaptation, masking, and patching (our full proposed method). 

For all five defect types, each step of the ablation resulted in incremental improvements in P-AUROC, demonstrating the benefits of fine-tuning, masking, patch-based processing, and low-rank adaptation in enhancing defect detection accuracy for surgical instruments. 


\begin{table}[ht]
\centering
\small
\resizebox{\columnwidth}{!}{%
\setlength{\tabcolsep}{3pt} 
\begin{tabular}{cccc|ccccc}
\hline
\multicolumn{4}{c|}{Components} & \multicolumn{5}{c}{P-AUROC} \\
Finetuned & Mask & Patch & LoRA & Blood & Pores & Cut & Corr. & Crack \\
\hline
 & & & & 0.802 & 0.480 & 0.523 & 0.429 & 0.549 \\ 
\checkmark & & & & 0.898 & 0.707 & 0.782 & 0.904 & 0.747 \\ 
\checkmark & \checkmark & & & 0.930 & 0.883 & 0.862 & 0.938 & 0.883 \\ 
\checkmark & \checkmark & \checkmark & & 0.951 & 0.920 & 0.899 & 0.941 & 0.899 \\ 
\checkmark & \checkmark & \checkmark & \checkmark & 0.954 & 0.927 & 0.902 & 0.942& 0.933 \\ 
\hline
\end{tabular}%
}
\caption{Ablation study of our method components using Dinomaly.}
\label{tab:ablation}
\end{table}
\section{Conclusion}
\label{sec:conclusion}

In this work, we addressed the challenge of applying general-purpose defect detection methods to surgical instrument inspection, a domain where direct transfer often fails due to textured backgrounds, subtle defects, and domain shift. We presented an approach that adapts existing methods through background masking, patch-based analysis, and low-rank domain adaptation. These adaptations make current defect detection techniques more effective and reliable for medical instrument quality assurance. Looking forward, this approach paves the way for safer, automated surgical instrument inspection and can be extended to additional instrument types, real-time analysis, and integration with hospital workflows to further enhance surgical safety and operational efficiency.


{
    \small
    \bibliographystyle{IEEEbib}
    \bibliography{refs}
}

\end{document}